\documentclass[preprint,12pt]{elsarticle}

\usepackage{amssymb}
\usepackage{multirow}
\usepackage{booktabs}
\usepackage{mathrsfs}
\usepackage{longtable}
\usepackage{lscape}
\usepackage{tabularx}
\usepackage{epsfig}
\usepackage{colortbl}
\usepackage{slashbox}
\usepackage{dcolumn,longtable,hhline,colortbl}
\usepackage[table]{xcolor}
\usepackage[rotateright]{rotating}
\usepackage{multirow}
\usepackage{morefloats}
\usepackage{amsthm}

\newdefinition{definition}{Definition}
\journal{The Scientific World Journal}

\begin{document}

\begin{frontmatter}



\title{Multiscale probability transformation of basic probability assignment}


\author[swu]{Meizhu Li}
\author[swu]{Qi Zhang}
\author[swu,vu1,vu2]{Yong Deng\corref{cor}}
\ead{ydeng@swu.edu.cn; yongdeng@nwpu.edu.cn}

\cortext[cor]{Corresponding author: Yong Deng, School of Computer
and Information Science, Southwest University, Chongqing 400715,
China. TEL: +86-23-68254555}

\address[swu]{School of Computer and Information Science, Southwest University, Chongqing, 400715, China}
\address[vu1]{School of Automation, Northwestern Polytechnical University, Xi¡¯an, Shanxi, 710072, China}
\address[vu2]{School of Engineering, Vanderbilt University, Nashville, TN, 37235, USA}

\begin{abstract}
Decision making is still an open issue in the application of Dempster-Shafer evidence theory. A lot of works have been presented for it. In the transferable belief model (TBM), pignistic probabilities based on the basic probability assignments are used for decision making. In this paper, multiscale probability transformation of basic probability assignment based on the belief function and the plausibility function is proposed, which is a generalization of the pignistic probability transformation. In the multiscale probability function, a factor $q$ based on the Tsallis entropy is used to make the multiscale probabilities diversified. An example is shown that the multiscale probability transformation is more reasonable in the decision making.

\end{abstract}

\end{frontmatter}

\section{Introduction}\label{Introduction}
Since first proposed by Dempster \cite{dempster1967upper}, and then developed by Shafer \cite{shafer1976mathematical}, the Dempster-Shafer theory of evidence, which is also called Dempster-Shafer theory or evidence theory, has been paid much attentions for a long time and continually attracted growing interests. Even as a theory of reasoning under the uncertain environment, Dempster-Shafer theory has an advantage of directly expressing the \lq \lq uncertainty \rq \rq by assigning the probability to the subsets of the set composed of multiple objects, rather than to each of the individual objects, so it has been widely used in many fields \cite{bloch1996some,srivastava2003applications,cuzzolin2008geometric,masson2008ecm,denoeux2011maximum,Dengnew2011,denoeux2013maximum,yang2013discounted,yang2013novel,wei2013identifying,liu2013evidential,deng2014supplier}.

Due to improve the Dempster-Shafer theory of evidence, many studies have been devoted for combination rule of evidence \cite{yager1996aggregation,gebhardt1998parallel,yang2013discounted,lefevre2013preserve,yang2013evidential}, confliction problem \cite{yager1987dempster,lefevre2002belief,liu2006analyzing,schubert2011conflict,tchamova2012behavior}, generation of mass function \cite{bastian2010universal,cappellari2012systematic,liu2013belief,burger2013randomly,liu2014credal}, uncertain measure of evidence \cite{klir1991generalized,bachmann2010uncertainty,bronevich2010measures,baker2012measuring}, and so on \cite{couso2010independence,limbourg2010uncertainty,jirouvsek2011compositional,luo2012agent,karahan2013persistence,mao2014model,zhang2014response}. One open issue of evidence theory is the decision making based on the basic probability assignments, many works have been done to construct a reasonable model for the decision making \cite{smets1994transferable,smets2005decision,cobb2006plausibility,daniel2006transformations,merigo2011decision,nusrat2013descriptive}.

In the transferable belief model (TBM) \cite{smets1994transferable}, pignistic probabilities are used for decision making. The transferable belief model is presented to represent quantified beliefs based on belief functions. TBM was constructed by two levels. The credal level where beliefs are entertained and quantified by belief functions. The pignistic level where beliefs can be used to make decisions and are quantified by probability functions. The main idea of the pignistic probability transformation is to transform the multi-elements subsets into singleton subsets by an average method. Though the pignistic probability transformation is widely used, it can not describe the unknown for the multi-elements subsets. Hense, a generalization of the pignistic probability transformation called multiscale probability transformation of basic probability assignment is proposed in this paper, which is based on the belief function and the plausibility function. The proposed function can be calculated with the difference between the belief function and the plausibility function, we call it multiscale probability function and denote it as a function $MulP$. In the multiscale probability function, a factor $q$ based on the Tsallis entropy \cite{tsallis1988possible} is used to make the multiscale probabilities diversified. When the value of $q$ equals to 0, the proposed multiscale probability transformation can be degenerated as the pignistic probability transformation.

The rest of this paper is organized as follows. Section \ref{Preliminaries} introduces some basic Preliminaries about the Dempster-Shafer theory and the pignistic probability transformation. In section \ref{Proposed function} the multiscale probability transformation is presented. Section \ref{Case study} uses an example to illustrate the effectiveness of the multiscale probability transformation. Conclusion is given in Section \ref{Conclusion}.

\section{Preliminaries}\label{Preliminaries}

\subsection{Dempster-Shafer theory of evidence}

Dempster-Shafer theory of evidence \cite{dempster1967upper,shafer1976mathematical}, also called Dempster-Shafer theory or evidence theory, is used to deal with uncertain information. As an effective theory of evidential reasoning, Dempster-Shafer theory has an advantage of directly expressing various uncertainties. This theory needs weaker conditions than bayesian theory of probability, so it is often regarded as an extension of the bayesian theory. For completeness of the explanation, a few basic concepts are introduced as follows.

\begin{definition}
Let $\Omega$ be a set of mutually exclusive and collectively
exhaustive, indicted by
\begin{equation}\label{q_1}
\Omega  = \{ E_1 ,E_2 , \cdots ,E_i , \cdots ,E_N \}
\end{equation}
The set $\Omega$ is called frame of discernment. The power set of
$\Omega$ is indicated by $2^\Omega$, where
\begin{equation}\label{q_2}
2^\Omega   = \{ \emptyset ,\{ E_1 \} , \cdots ,\{ E_N \} ,\{ E_1
,E_2 \} , \cdots ,\{ E_1 ,E_2 , \cdots ,E_i \} , \cdots ,\Omega \}
\end{equation}
If $A \in 2^\Omega$, $A$ is called a proposition.
\end{definition}

\begin{definition}
For a frame of discernment $\Omega$,  a mass function is a mapping
$m$ from  $2^\Omega$ to $[0,1]$, formally defined by:
\begin{equation}\label{q_3}
m: \quad 2^\Omega \to [0,1]
\end{equation}
which satisfies the following condition:
\begin{eqnarray}\label{q_4}
m(\emptyset ) = 0 \quad and \quad \sum\limits_{A \in 2^\Omega }
{m(A) = 1}
\end{eqnarray}
\end{definition}

In Dempster-Shafer theory, a mass function is also called a basic
probability assignment (BPA). If $m(A) > 0$, $A$ is called a focal
element, the union of all focal elements is called the core of the
mass function.

\begin{definition}
For a proposition $A \subseteq \Omega$, the belief function
$Bel:\;2^\Omega   \to [0,1]$ is defined as
\begin{equation}\label{q_5}
Bel(A) = \sum\limits_{B \subseteq A} {m(B)}
\end{equation}
The plausibility function $Pl:\;2^\Omega   \to [0,1]$ is defined
as
\begin{equation}\label{q_6}
Pl(A) = 1 - Bel(\bar A) = \sum\limits_{B \cap A \ne \emptyset }
{m(B)}
\end{equation}
where $\bar A = \Omega  - A$.
\end{definition}

Obviously, $Pl(A) \ge Bel(A)$, these functions $Bel$ and $Pl$ are
the lower limit function and upper limit function of proposition
$A$, respectively.

\subsection{Pignistic probability transformation}

In the transferable belief model (TBM) \cite{smets1994transferable}, pignistic probabilities are used for decision making. The definition of the pignistic probability transformation is shown as follows.

\begin{definition}
Let $m$ be a BPA on the frame of discernment $\Omega$. Its associated pignistic probability function $Bet{P_m}:\Omega   \to [0,1]$ is defined as:

\begin{equation}\label{q_7}
Bet{P_m}(\omega ) = \sum\limits_{A \subseteq P(\Omega ),\omega  \in A} {\frac{1}{{\left| A \right|}}\frac{{m(A)}}{{1 - m(\phi )}}} ,\  \   m(\phi ) \ne 1
\end{equation}

where ${\left| W \right|}$  is the cardinality of subset A. The process of pignistic probability transformation(PPT) is that basic probability assignment transferred to probability distribution. Therefore, the pignistic betting distance can be easily obtained by PPT.

\end{definition}

\section{Multiscale probability transformation of basic probability assignment}\label{Proposed function}

In the transferable belief model (TBM) \cite{smets1994transferable}, pignistic probabilities are used for decision making. The transferable belief model is presented to represent quantified beliefs based on belief functions. The main idea of the pignistic probability transformation is to transform the multi-elements subsets into singleton subsets by an average method. Though the pignistic probability transformation is widely used, it is not reasonable in the Example 1.

\textbf{Example 1.}
Suppose there is a frame of discernment of {a, b, c}, the BPA is given as follows.

${m}(\{ a\} ) = 0.2$, ${m}(\{ b\} ) = 0.7$, ${m}(\{ b,c\} ) = 0.05$, ${m}(\{ a,b,c\} ) = 0.05$.

In the pignistic probability transformation, for ${m}(\{ a,b,c\} ) = 0.05$, the result will be $a=b=c=0.05/3$. Actually it is not reasonable, ${m}(\{ a,b,c\} ) = 0.05$ means the sensor can not judge the target belongs to which classes, it represents a meaning of \lq \lq unknown \rq \rq. In other word, only according to ${m}(\{ a,b,c\} ) = 0.05$, nothing can be obtained except \lq \lq unknown \rq \rq. In this situation, average is used in the pignistic probability transformation, which is one of the methods to solve the problem. Compared with the average, weighted average is more reasonable in many situations. In this paper, the weighted average is represented by the difference between the belief function and the plausibility function, whose definition is shown as follows.

\begin{definition}
Let $m$ be a BPA on the frame of discernment $\Omega$. The difference function ${d_m}$ is defined as:
\begin{equation}\label{q_8}
{d_m}(\omega ) = Pl(\omega ) - Bel(\omega ),\  \  \omega  \in \Omega
\end{equation}
\end{definition}

\begin{definition}
The weight is defined as:
\begin{equation}\label{q_9}
{Weight_m}(\omega ) = \frac{{{d_m}(\omega )}}{{\sum\limits_{\alpha  \in A}^{\left| A \right|} {{d_m}(\alpha )} }}, \  \  \omega  \in A,A \subseteq P(\Omega )
\end{equation}
\end{definition}

Based on the weighted average idea, a factor $q$, which is proposed in the Tsallis entropy \cite{tsallis1988possible}, is used to highlight the weights. Thus, the definition of multiscale probability function $MulP$ is shown as follows.

\begin{definition}
Let $m$ be a BPA on the frame of discernment $\Omega$. Its associated multiscale probability function $Mul{P_m}:\Omega    \to [0,1]$ on $\Omega$ is defined as:

\begin{equation}\label{q_10}
Mul{P_m}(\omega ) = \sum\limits_{A \subseteq P(\Omega ),\omega  \in A} {\left( {\frac{{{{(Pl(\omega ) - Bel(\omega ))}^q}}}{{\sum\limits_{\alpha  \in A}^{\left| A \right|} {{{(Pl(\alpha ) - Bel(\alpha ))}^q}} }}\frac{{m(A)}}{{1 - m(\phi )}}} \right)} , \  \  m(\phi ) \ne 1
\end{equation}

where ${\left| W \right|}$  is the cardinality of subset A. $q$ is a factor based on the Tsallis entropy to amend the proportion of the interval. The transformation between $m$ and $Mul{P_m }$ is called the multiscale probability transformation.
\end{definition}

Actually, the part of the Eq. \ref{q_10} ${\frac{{{{\left( {Pl(\omega ) - Bel(\omega )} \right)}^q}}}{{\sum\limits_{\alpha  \in W}^{\left| W \right|} {{{\left( {Pl(\alpha ) - Bel(\alpha )} \right)}^q}} }}}$ denotes the weight of element $\omega$ based on normalization, which is replaced the averaged ${\frac{1}{{\left| W \right|}}}$ in the pignistic probability function.

\textbf{Theorem 3.1:} Let $m$ be a BPA on the frame of discernment $\Omega$. Its associated multiscale probability $Mul{P_m}$ on $\Omega$ is degenerated as the pignistic probability $Bet{P_m}$ when $q$ equals to 0.

\textbf{Proof:} When $q$ equals to 0, ${{{\left( {Pl(\omega ) - Bel(\omega )} \right)}^q}}$ equals to 1, the multiscale probability function will be calculated as follows:
\begin{equation}\label{q_11}
Mul{P_m }(\omega ) = \sum\limits_{A \subseteq \Omega ,\omega  \in A} {\left( {\frac{1}{{\sum\limits_{\alpha  \in A}^{\left| A \right|} 1 }} \cdot \frac{{{m^\Omega }(W)}}{{(1 - {m^\Omega }(\phi ))}}} \right)} ,\forall \omega  \in \Omega
\end{equation}
Then, it can obtain:
\begin{equation}\label{q_12}
Mul{P_m }(\omega ) = \sum\limits_{A \subseteq \Omega ,\omega  \in A} {\left( {\frac{1}{{\left| A\right|}} \cdot \frac{{{m^\Omega }(A)}}{{(1 - {m^\Omega }(\phi ))}}} \right)} ,\forall \omega  \in \Omega
\end{equation}

From Eq. \ref{q_11} and Eq. \ref{q_12}, we can see that when the value of $q$ equals to 0, the proposed multiscale probability function can be degenerated as the pignistic probability function.

\textbf{Theorem 3.2:} Let $m$ be a BPA on the frame of discernment $\Omega$. If the belief function equals to the plausibility function, its associated multiscale probability $Mul{P_m}$ on is degenerated as the pignistic probability $Bet{P_m}$.

\textbf{Proof:} Given a BPA $m$ on the frame of discernment $\Omega$, for each $\omega  \in \Omega $, when the belief function equals to the plausibility function, namely $Bel(\omega ) = Pl(\omega )$, the bel is a probability distribution P \cite{smets1994transferable}, then MulP is equal to BetP.

For example, let $\Omega$ be a frame of discernment and $\Omega  = \{ a,b,c\} $, if it satisfies with $Bel(a) = Pl(a)$, $Bel(b) = Pl(b)$, $Bel(c) = Pl(c)$, the BPA on the frame must be satisfied with $m(a)+m(b)+m(c)=1$. In this situation, the multiscale probability will be degenerated as the pignistic probability.

\textbf{Corollary:} If bel is a probability distribution P, then MulP is equal to P.

\textbf{Theorem 3.3:} Let $m$ be a BPA on the frame of discernment $\Omega={a,b,c}$. If the differences between the belief function and the plausibility function is the same, the multiscale probability transformation can be degenerated as the pignistic probability transformation.

\textbf{Proof:} The same as the proof of theorem 3.1.

An illustrative example is given to show the calculation of the multiscale probability transformation step by step.

\textbf{Example 2.}
Let $\Omega$ be a frame of discernment with 3 elements. We use $a$, $b$, $c$ to denote element 1, element 2, and element 3 in the frame. One body of BPA is given as follows:

${m}(\{ a\} ) = 0.2$,

${m}(\{ b\} ) = 0.3$,

${m}(\{ c\} ) = 0.1$,

${m}(\{ a,b\} ) = 0.1$,

${m}(\{ a,b,c\} ) = 0.3$.

\textbf{Step 1}
Based on Eq. \ref{q_5} and Eq. \ref{q_6}, the values of the belief function and the plausibility function of elements $a$, $b$, $c$ can be obtained as follows:

$Bel(a)=0.2$, $Pl(a)=0.6$,

$Bel(b)=0.3$, $Pl(b)=0.7$,

$Bel(c)=0.1$, $Pl(c)=0.4$.

\textbf{Step 2}
Calculate the difference between the belief function and the plausibility function:

$d_m(a) = Pl(a)-Bel(a)=0.6-0.2=0.4$,

$d_m(b) = Pl(b)-Bel(b)=0.7-0.3=0.4$,

$d_m(c) = Pl(c)-Bel(c)=0.4-0.1=0.3$.

\textbf{Step 3}
Calculate the weight of each element in $\Omega$. Assumed that the value of $q$ equals to 1.

When $A = \{ a,b\} $,

$Weight_m(a)=\frac{{\left( {Pl(a) - Bel(a)} \right)}}{{\sum\limits_{\alpha  \in W}^{\left| W \right|} {\left( {Pl(\alpha ) - Bel(\alpha )} \right)} }}=0.4/(0.4+0.4)=0.5$,

$Weight_m(b)=\frac{{\left( {Pl(b) - Bel(b)} \right)}}{{\sum\limits_{\alpha  \in W}^{\left| W \right|} {\left( {Pl(\alpha ) - Bel(\alpha )} \right)} }}=0.4/(0.4+0.4)=0.5$.

When $W = \{ a,b,c\}$,

$Weight_m(a)=\frac{{\left( {Pl(a) - Bel(a)} \right)}}{{\sum\limits_{\alpha  \in W}^{\left| W \right|} {\left( {Pl(\alpha ) - Bel(\alpha )} \right)} }}=0.4/(0.4+0.4+0.3)=0.364$,

$Weight_m(b)=\frac{{\left( {Pl(b) - Bel(b)} \right)}}{{\sum\limits_{\alpha  \in W}^{\left| W \right|} {\left( {Pl(\alpha ) - Bel(\alpha )} \right)} }}=0.4/(0.4+0.4+0.3)=0.364$,

$Weight_m(c)=\frac{{\left( {Pl(c) - Bel(c)} \right)}}{{\sum\limits_{\alpha  \in W}^{\left| W \right|} {\left( {Pl(\alpha ) - Bel(\alpha )} \right)} }}=0.3/(0.4+0.4+0.3)=0.272$.

\textbf{Step 4}
The value of the multiscale probability function can be obtained based on above steps.

$Mul{P_m }(a) = 0.2+0.1*0.5+0.3*0.364=0.3592 $,

$Mul{P_m }(b) = 0.3+0.1*0.5+0.3*0.364=0.4592 $,

$Mul{P_m }(c) = 0.1+0.3*0.272=0.1816 $.

\section{Case study}\label{Case study}
In this section, an illustrative example is given to show the effection of the multiscale probability function when the value of $q$ changes.

\textbf{Example 3.} Let $\Omega$ be a frame of discernment with 3 elements, namely $\Omega  = \{ a,b,c\}$.

Given one body of BPAs:

${m}(\{ a\} ) = 0.3$,

${m}(\{ b\} ) = 0.1$,

${m}(\{ a,b\} ) = 0.1$,

${m}(\{ a,c\} ) = 0.2$,

${m}(\{ a,b,c\} ) = 0.3$.

Based on the pignistic probability transformation, the results of the pignistic probability function is shown as follows:

$Bet{P_m }(a )=0.55$,

$Bet{P_m }(b )=0.25$,

$Bet{P_m }(c )=0.20$.

According to the proposed function in this paper, the results of the multiscale probability function can be obtained through the follow steps.

Firstly, the values of belief function and the plausibility function can be obtained as follows:

$Bel(a)=0.3$, $Pl(a)=0.9$,

$Bel(b)=0.1$, $Pl(b)=0.5$,

$Bel(c)=0$, $Pl(c)=0.5$.

Then, the differences between the belief functions and the plausibility functions can be calculated:

$d_m(a)=Pl(a)-Bel(a)=0.6-0.2=0.6$,

$d_m(b)=Pl(b)-Bel(b)=0.7-0.3=0.4$,

$d_m(c)=Pl(c)-Bel(c)=0.4-0.1=0.5$.

\begin{figure}[htbp]
\begin{center}
\psfig{file=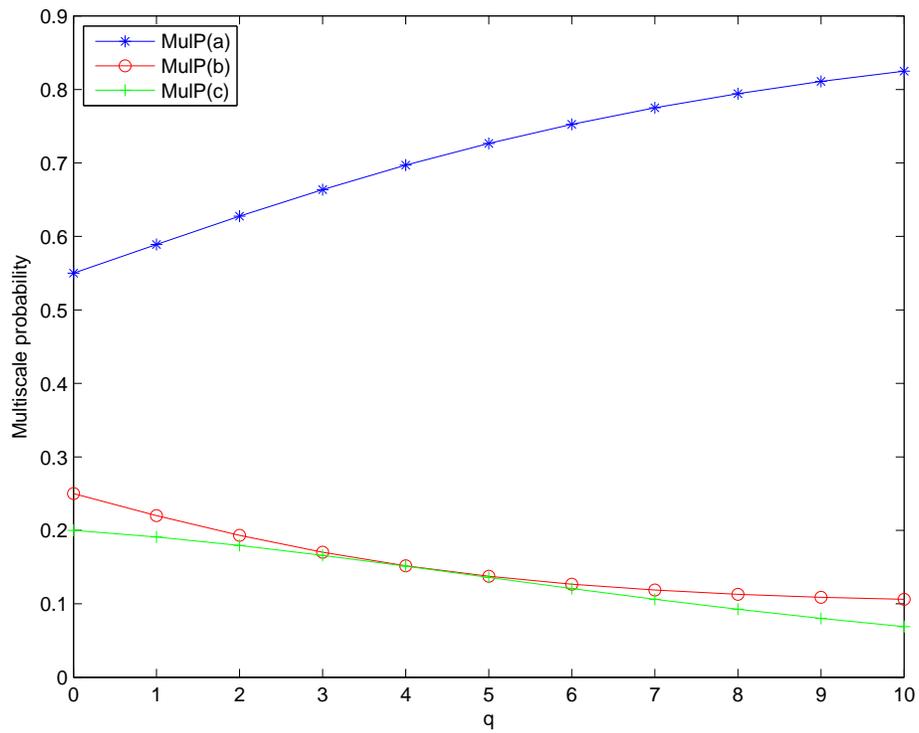,scale=0.75} \caption{The values of multiscale probability function when the values of $q$ changes.} \label{q}
\end{center}
\end{figure}

Based on the definition of the multiscale probability transformation, the values of $Mul{P_m }$ can be obtained. There are 20 cases where the values of $q$ starting from Case 1 with $q=0$ and ending with Case 11 when $q=10$ as shown in Table \ref{tab_1}. The values of $Mul{P_m }$ for these 20 cases is detailed in Table \ref{tab_1} and graphically illustrated in Fig. \ref{q}.

\begin{table}[htbp]
\caption{The values of multiscale probability function when the values of $q$ changes.}
\label{tab_1}
\begin{center}
\begin{tabular}{lcccccccccccccc}
\toprule Cases   &   &   &   & $MulP(a)$&   &   &   &  $MulP(b)$&   &   &   &  $MulP(c)$ \\
\midrule
q=0  &   &   &   &  0.5500  &   &   &   &  0.2500&   &   &   &  0.2000     \\
q=1  &   &   &   &  0.5891  &   &   &   &  0.2200&   &   &   &  0.1909     \\
q=2  &   &   &   &  0.6275  &   &   &   &  0.1931&   &   &   &  0.1794     \\
q=3  &   &   &   &  0.6638  &   &   &   &  0.1703&   &   &   &  0.1659     \\
q=4  &   &   &   &  0.5970  &   &   &   &  0.1518&   &   &   &  0.1512     \\
q=5  &   &   &   &  0.7267  &   &   &   &  0.1374&   &   &   &  0.1360     \\
q=6  &   &   &   &  0.7526  &   &   &   &  0.1266&   &   &   &  0.1208     \\
q=7  &   &   &   &  0.7751  &   &   &   &  0.1187&   &   &   &  0.1062     \\
q=8  &   &   &   &  0.7944  &   &   &   &  0.1130&   &   &   &  0.0926     \\
q=9  &   &   &   &  0.8109  &   &   &   &  0.1089&   &   &   &  0.0801      \\
q=10 &   &   &   &  0.8250  &   &   &   &  0.1061&   &   &   &  0.0689      \\
\bottomrule
\end{tabular}
\end{center}
\end{table}

According to the Table \ref{tab_1} and the Fig. \ref{q}, on one hand, when the value of $q$ increased, the probability of the element which has larger weight is increased, and the probability of the element which has smaller weight is decreased. For example, the element $a$ starting with probability 0.5500, and ending with probability 0.8250. The element $b$ starting with probability 0.2500, and ending with probability 0.1061.

On the other hand, according to the Table \ref{tab_1}, the option ranking of the the values of $Mul{P_m }$ can be obtained. It is starting with $Mul{P_m }(a) \succ Mul{P_m }(b) \succ Mul{P_m }(c)$, and ending with $Mul{P_m }(a) \succ Mul{P_m }(c) \succ Mul{P_m }(b)$. It is mainly because $Mul{P_m }$ is impact of the values of $q$. This principle makes the multiscale probability function has the ability to highlight the proportion of each element in the frame of discernment.

Note that when the value of $q$ equals to 0, the values of pignistic probability $Bet{P_m }$ is the same as the values of multiscale probability $Mul{P_m }$, which is proposed in this paper. In other word, the multiscale probability function is a generalization of the pignistic probability function.

\section{Conclusion}\label{Conclusion}
In the transferable belief model(TBM), pignistic probabilities are used for decision making. In this paper, a multiscale probability transformation of basic probability assignment based on the belief function and the plausibility function, which is a generalization of the pignistic probability transformation is proposed. In the multiscale probability function, a factor $q$ is proposed to make the multiscale probability function has the ability to highlight the proportion of each element in the frame of discernment. When the value of $q$ equals to 0, the multiscale probability transformation can be degenerated as the pignistic probability transformation. An illustrative case is provided to demonstrate the effectiveness of the multiscale probability transformation.

\section*{Acknowledgements}
The work is partially supported by National Natural Science Foundation of China (Grant No. 61174022), Specialized Research Fund for the Doctoral Program of Higher Education (Grant No. 20131102130002), R\&D Program of China (2012BAH07B01), National High Technology Research and Development Program of China (863 Program) (Grant No. 2013AA013801), the open funding project of State Key Laboratory of Virtual Reality Technology and Systems, Beihang University (Grant No.BUAA-VR-14KF-02).

\section*{Conflict of interests}
The authors declare that there is no conflict of interests regarding the publication of this article.

%\appendix
%
%\section{1}

%% References


\bibliographystyle{elsarticle-num}
\bibliography{reference}

\begin{thebibliography}{10}
\expandafter\ifx\csname url\endcsname\relax
  \def\url#1{\texttt{#1}}\fi
\expandafter\ifx\csname urlprefix\endcsname\relax\def\urlprefix{URL }\fi
\expandafter\ifx\csname href\endcsname\relax
  \def\href#1#2{#2} \def\path#1{#1}\fi

\bibitem{dempster1967upper}
A.~P. Dempster, Upper and lower probabilities induced by a multivalued mapping,
  The annals of mathematical statistics 38~(2) (1967) 325--339.

\bibitem{shafer1976mathematical}
G.~Shafer, A mathematical theory of evidence, Vol.~1, Princeton university
  press Princeton, 1976.

\bibitem{bloch1996some}
I.~Bloch, Some aspects of dempster-shafer evidence theory for classification of
  multi-modality medical images taking partial volume effect into account,
  Pattern Recognition Letters 17~(8) (1996) 905--919.

\bibitem{srivastava2003applications}
R.~P. Srivastava, L.~Liu, Applications of belief functions in business
  decisions: A review, Information Systems Frontiers 5~(4) (2003) 359--378.

\bibitem{cuzzolin2008geometric}
F.~Cuzzolin, A geometric approach to the theory of evidence, Systems, Man, and
  Cybernetics, Part C: Applications and Reviews, IEEE Transactions on 38~(4)
  (2008) 522--534.

\bibitem{masson2008ecm}
M.-H. Masson, T.~Denoeux, Ecm: An evidential version of the fuzzy c-means
  algorithm, Pattern Recognition 41~(4) (2008) 1384--1397.

\bibitem{denoeux2011maximum}
T.~Den{\oe}ux, Maximum likelihood estimation from fuzzy data using the em
  algorithm, Fuzzy sets and systems 183~(1) (2011) 72--91.

\bibitem{Dengnew2011}
Y.~Deng, F.~T. Chan, A new fuzzy dempster mcdm method and its application in
  supplier selection, Expert Systems with Applications 38 (2011) 9854--9861.

\bibitem{denoeux2013maximum}
T.~Denoeux, Maximum likelihood estimation from uncertain data in the belief
  function framework, Knowledge and Data Engineering, IEEE Transactions on
  25~(1) (2013) 119--130.

\bibitem{yang2013discounted}
Y.~Yang, D.~Han, C.~Han, Discounted combination of unreliable evidence using
  degree of disagreement, International Journal of Approximate Reasoning 54~(8)
  (2013) 1197--1216.

\bibitem{yang2013novel}
Y.~Yang, D.~Han, C.~Han, F.~Cao, A novel approximation of basic probability
  assignment based on rank-level fusion, Chinese Journal of Aeronautics 26~(4)
  (2013) 993--999.

\bibitem{wei2013identifying}
D.~Wei, X.~Deng, X.~Zhang, Y.~Deng, S.~Mahadevan, Identifying influential nodes
  in weighted networks based on evidence theory, Physica A: Statistical
  Mechanics and its Applications 392~(10) (2013) 2564--2575.

\bibitem{liu2013evidential}
Z.-g. Liu, Q.~Pan, J.~Dezert, Evidential classifier for imprecise data based on
  belief functions, Knowledge-Based Systems 52 (2013) 246--257.

\bibitem{deng2014supplier}
X.~Deng, Y.~Hu, Y.~Deng, S.~Mahadevan, Supplier selection using ahp methodology
  extended by d numbers, Expert Systems with Applications 41~(1) (2014)
  156--167.

\bibitem{yager1996aggregation}
R.~R. Yager, On the aggregation of prioritized belief structures, Systems, Man
  and Cybernetics, Part A: Systems and Humans, IEEE Transactions on 26~(6)
  (1996) 708--717.

\bibitem{gebhardt1998parallel}
J.~Gebhardt, R.~Kruse, Parallel combination of information sources, in: Belief
  Change, Springer, 1998, pp. 393--439.

\bibitem{lefevre2013preserve}
E.~Lef{\`e}vre, Z.~Elouedi, How to preserve the conflict as an alarm in the
  combination of belief functions?, Decision Support Systems 56 (2013)
  326--333.

\bibitem{yang2013evidential}
J.-B. Yang, D.-L. Xu, Evidential reasoning rule for evidence combination,
  Artificial Intelligence 205 (2013) 1--29.

\bibitem{yager1987dempster}
R.~R. Yager, On the dempster-shafer framework and new combination rules,
  Information sciences 41~(2) (1987) 93--137.

\bibitem{lefevre2002belief}
E.~Lefevre, O.~Colot, P.~Vannoorenberghe, Belief function combination and
  conflict management, Information fusion 3~(2) (2002) 149--162.

\bibitem{liu2006analyzing}
W.~Liu, Analyzing the degree of conflict among belief functions, Artificial
  Intelligence 170~(11) (2006) 909--924.

\bibitem{schubert2011conflict}
J.~Schubert, Conflict management in dempster--shafer theory using the degree of
  falsity, International Journal of Approximate Reasoning 52~(3) (2011)
  449--460.

\bibitem{tchamova2012behavior}
A.~Tchamova, J.~Dezert, On the behavior of dempster's rule of combination and
  the foundations of dempster-shafer theory, in: Intelligent Systems (IS), 2012
  6th IEEE International Conference, IEEE, 2012, pp. 108--113.

\bibitem{bastian2010universal}
N.~Bastian, K.~R. Covey, M.~R. Meyer, A universal stellar initial mass
  function? a critical look at variations, arXiv preprint arXiv:1001.2965.

\bibitem{cappellari2012systematic}
M.~Cappellari, R.~M. McDermid, K.~Alatalo, L.~Blitz, M.~Bois, F.~Bournaud,
  M.~Bureau, A.~F. Crocker, R.~L. Davies, T.~A. Davis, et~al., Systematic
  variation of the stellar initial mass function in early-type galaxies, Nature
  484~(7395) (2012) 485--488.

\bibitem{liu2013belief}
Z.-g. Liu, Q.~Pan, J.~Dezert, A belief classification rule for imprecise data,
  Applied Intelligence (2013) 1--15.

\bibitem{burger2013randomly}
T.~Burger, S.~Destercke, How to randomly generate mass functions, International
  Journal of Uncertainty, Fuzziness and Knowledge-Based Systems 21~(05) (2013)
  645--673.

\bibitem{liu2014credal}
Z.-g. Liu, Q.~Pan, J.~Dezert, G.~Mercier, Credal classification rule for
  uncertain data based on belief functions, Pattern Recognition 47~(7) (2014)
  2532--2541.

\bibitem{klir1991generalized}
G.~J. Klir, Generalized information theory, Fuzzy sets and systems 40~(1)
  (1991) 127--142.

\bibitem{bachmann2010uncertainty}
R.~Bachmann, S.~Elstner, E.~R. Sims, Uncertainty and economic activity:
  Evidence from business survey data, Tech. rep., National Bureau of Economic
  Research (2010).

\bibitem{bronevich2010measures}
A.~Bronevich, G.~J. Klir, Measures of uncertainty for imprecise probabilities:
  An axiomatic approach, International journal of approximate reasoning 51~(4)
  (2010) 365--390.

\bibitem{baker2012measuring}
S.~R. Baker, N.~Bloom, S.~J. Davis, Measuring economic policy uncertainty,
  policyuncertainy. com.

\bibitem{couso2010independence}
I.~Couso, S.~Moral, Independence concepts in evidence theory, International
  Journal of Approximate Reasoning 51~(7) (2010) 748--758.

\bibitem{limbourg2010uncertainty}
P.~Limbourg, E.~De~Rocquigny, Uncertainty analysis using evidence
  theory--confronting level-1 and level-2 approaches with data availability and
  computational constraints, Reliability Engineering \& System Safety 95~(5)
  (2010) 550--564.

\bibitem{jirouvsek2011compositional}
R.~Jirou{\v{s}}ek, J.~Vejnarov{\'a}, Compositional models and conditional
  independence in evidence theory, International Journal of Approximate
  Reasoning 52~(3) (2011) 316--334.

\bibitem{luo2012agent}
H.~Luo, S.-l. Yang, X.-j. Hu, X.-x. Hu, Agent oriented intelligent fault
  diagnosis system using evidence theory, Expert Systems with Applications
  39~(3) (2012) 2524--2531.

\bibitem{karahan2013persistence}
F.~Karahan, S.~Ozkan, On the persistence of income shocks over the life cycle:
  Evidence, theory, and implications, Review of Economic Dynamics 16~(3) (2013)
  452--476.

\bibitem{mao2014model}
S.~Mao, Z.~Zou, Y.~Xue, Y.~Li, A model based on the coupled rules of evidence
  theory used in multiple objective decisions, in: 2014 International
  Conference on Global Economy, Finance and Humanities Research (GEFHR 2014),
  Atlantis Press, 2014.

\bibitem{zhang2014response}
Z.~Zhang, C.~Jiang, X.~Han, D.~Hu, S.~Yu, A response surface approach for
  structural reliability analysis using evidence theory, Advances in
  Engineering Software 69 (2014) 37--45.

\bibitem{smets1994transferable}
P.~Smets, R.~Kennes, The transferable belief model, Artificial intelligence
  66~(2) (1994) 191--234.

\bibitem{smets2005decision}
P.~Smets, Decision making in the tbm: the necessity of the pignistic
  transformation, International Journal of Approximate Reasoning 38~(2) (2005)
  133--147.

\bibitem{cobb2006plausibility}
B.~R. Cobb, P.~P. Shenoy, On the plausibility transformation method for
  translating belief function models to probability models, International
  Journal of Approximate Reasoning 41~(3) (2006) 314--330.

\bibitem{daniel2006transformations}
M.~Daniel, On transformations of belief functions to probabilities,
  International Journal of Intelligent Systems 21~(3) (2006) 261--282.

\bibitem{merigo2011decision}
J.~M. Merig{\'o}, M.~Casanovas, Decision making with dempster-shafer theory
  using fuzzy induced aggregation operators, in: Recent developments in the
  ordered weighted averaging operators: Theory and practice, Springer, 2011,
  pp. 209--228.

\bibitem{nusrat2013descriptive}
E.~Nusrat, K.~Yamada, A descriptive decision-making model under uncertainty:
  combination of dempster-shafer theory and prospect theory, International
  Journal of Uncertainty, Fuzziness and Knowledge-Based Systems 21~(01) (2013)
  79--102.

\bibitem{tsallis1988possible}
C.~Tsallis, Possible generalization of {B}oltzmann-{G}ibbs statistics, Journal
  of Statistical Physics 52~(1-2) (1988) 479--487.

\end{thebibliography}

 \section*{Biographical Note}
Yong Deng received his Ph.D. degree from Shanghai Jiao Tong University, China, 2003. He is a full professor in School of Computer and Information Science of Southwest University, Chongqing, China and a visiting professor in School of Engineering of Vanderbilt University, TN, USA. Now he is a member of the editorial board for \emph{The Scientific World Journal}. So far, he has published more than 100 peer-reviewed journal. His research interests include uncertain information modeling, risk and reliability analysis, information fusion and optimization under uncertain environment.






% \end{thebibliography}

\end{document}